# Locally Weighted Naive Bayes

Eibe Frank, Mark Hall, and Bernhard Pfahringer
Department of Computer Science
University of Waikato
Hamilton, New Zealand
{eibe, mhall, bernhard}@cs.waikato.ac.nz

## Abstract

Despite its simplicity, the naive Bayes classifier has surprised machine learning researchers by exhibiting good performance on a variety of learning problems. Encouraged by these results, researchers have looked to overcome naive Bayes' primary weakness—attribute independence—and improve the performance of the algorithm. This paper presents a locally weighted version of naive Bayes that relaxes the independence assumption by learning local models at prediction time. Experimental results show that locally weighted naive Bayes rarely degrades accuracy compared to standard naive Bayes and, in many cases, improves accuracy dramatically. The main advantage of this method compared to other techniques for enhancing naive Bayes is its conceptual and computational simplicity.

## 1 Introduction

In principle, Bayes' theorem enables optimal prediction of the class label for a new instance given a vector of attribute values. Unfortunately, straightforward application of Bayes' theorem for machine learning is impractical because inevitably there is insufficient training data to obtain an accurate estimate of the full joint probability distribution. Some independence assumptions have to be made to make inference feasible. The naive Bayes approach takes this to the extreme by assuming that the attributes are statistically independent given the value of the class attribute. Although this assumption never holds in practice, naive Bayes performs surprisingly well in many classification problems. Furthermore, it is computationally efficient—training is linear in both the number of instances and attributes—and simple to implement.

Interest in the naive Bayes learning algorithm within machine learning circles can be attributed to Clark and Niblett's paper on the CN2 rule learner (Clark & Niblett, 1989). In this paper they included a simple Bayesian classifier (naive Bayes) as a "straw man" in their experimental evaluation and noted its good performance compared to more sophisticated learners. Although it has been explained how naive Bayes can work well in some cases where the attribute independence assumption is violated (Domingos & Pazzani, 1997) the fact remains that probability estimation is less accurate and performance degrades when attribute independence does not hold.

Various techniques have been developed to improve the performance of naive Bayes—many of them aimed at reducing the 'naivete' of the algorithm—while still retaining the desirable aspects of simplicity and computational efficiency. Zheng and Webb (2000) provide an excellent overview of work in this area. Most existing techniques involve restricted sub-classes of Bayesian networks, combine attribute selection with naive Bayes, or incorporate naive Bayes models into another type of classifier (such as a decision tree).

This paper presents a lazy approach to learning naive Bayes models. Like all lazy learning methods our approach simply stores the training data and defers the effort involved in learning until classification time. When called upon to classify a new instance, we construct a new naive Bayes model using a weighted set of training instances in the locale of the test instance. Local learning helps to mitigate the effects of attribute dependencies that may exist in the data as a whole and we expect this method to do well if there are no strong dependencies within the neighbourhood of the test instance. Because naive Bayes requires relatively little data for training, the neighbourhood can be kept small, thereby reducing the chance of encountering strong dependencies. In our implementation the size of the neighbourhood is chosen in a data-dependent fashion based on the distance of the $k$-th nearest-neighbour to the test instance. Our experimental results show that locally weighted naive Bayes is relatively insensitive to the choice of $k$. This makes it a very attractive alternative to the $k$-nearest neighbour algorithm, which requires fine-tuning of $k$ to achieve good results. Our results also show that locally weighted naive Bayes almost uniformly improves on standard naive Bayes.

This paper is structured as follows. In Section 2 we present our approach for enhancing naive Bayes by using locally weighted learning. Section 3 contains experimental results for two artificial domains and a collection of benchmark datasets, demonstrating that the predictive accuracy of naive Bayes can be improved by learning locally weighted models at prediction time. Section 4 discusses related work on enhancing the performance of naive Bayes. Section 5 summarizes the contributions made in this paper.

## 2 Locally weighted learning with naive Bayes

Our method for enhancing naive Bayes borrows from a technique originally proposed for estimating non-linear regression models (Cleveland, 1979), where a linear regression model is fit to the data based on a weighting function centered on the instance for which a prediction is to be generated. The resulting estimator is non-linear because the weighting function changes with every instance to be processed. In this paper we explore locally weighted learning for classification (Atkeson et al., 1997), which appears to have received little attention in the machine learning literature.

Loader (1999) and Hastie et al. (2001) discuss so-called "local likelihood" methods from a statistical perspective, including locally weighted linear logistic regression and locally weighted density estimation. Naive Bayes is an example of using density estimation for classification. Compared to logistic regression it has the advantage that it is linear in the number of attributes, making it much more computationally efficient in learning problems with many attributes.

We use naive Bayes in exactly the same way as linear regression is used in locally weighted linear regression: a local naive Bayes model is fit to a subset of the data that is in the neighbourhood of the instance whose class value is to be predicted (we will call this instance the "test instance"). The training instances in this neighbourhood are weighted, with less weight being assigned to instances that are further from the test instance. A classification is then obtained from the naive Bayes model taking the attribute values of the test instance as input.

The subsets of data used to train each locally weighted naive Bayes model are determined by a nearest neighbours algorithm. A user-specified parameter $k$ controls how many instances are used. This is implemented by using a weighting function with compact support, setting its width (or "bandwidth") to the distance of the $k$th nearest neighbour.

Let $d_i$ be the Euclidean distance to the $i$th nearest neighbour $x_i$. We assume that all attributes have been normalized to lie between zero and one before the distance is computed, and that nominal attributes have been binarized. Let $f$ be a weighting function with $f(y) = 0$ for all $y \geq 1$. We then set the weight $w_i$ of each instance $x_i$ to

$$w_i = f(d_i/d_k) \qquad (1)$$

This means that instance $x_k$ receives weight zero, all instances that are further away from the test instance also receive weight zero, and an instance identical to the test instance receives weight one.

Any monotonically decreasing function with the above property is a candidate weighting function. In our experiments we used a linear weighting function $f_{linear}$ defined as

$$f_{linear}(y) = 1 - y \qquad \text{for } y \in [0, 1] \qquad (2)$$

In other words, we let the weight decrease linearly with the distance.

Higher values for $k$ result in models that vary less in response to fluctuations in the data, while lower values for $k$ enable models to conform closer to the data. Too small a value for $k$ may result in models that fit noise in the data. Our experiments show that the method is not particularly sensitive to the choice of $k$ as long as it is not too small.

There is one caveat. In order to avoid the zero-frequency problem our implementation of naive Bayes uses the Laplace estimator to estimate the conditional probabilities for nominal attributes and this interacts with the weighting scheme. We found empirically that it is opportune to scale the weights so that the total weight of the instances used to generate the naive Bayes model is approximately $k$. Assume that there are $r$ training instances $x_i$ with $d_i \leq d_k$. Then the rescaled weights $w'_i$ are computed as follows:

$$w'_i = \frac{w_i \times r}{\sum_{q=0}^{n} w_q}, \qquad (3)$$

where $n$ is the total number of training instances.

Naive Bayes computes the posterior probability of class $c_l$ for a test instance with attribute values

$a_1, a_2, ..., a_m$ as follows:

$$p(c_l|a_1, a_2, ..., a_m) = \frac{p(c_l) \prod_{j=1}^{m} p(a_j|c_l)}{\sum_{q=1}^{o} \left[ p(c_q) \prod_{j=1}^{m} p(a_j|c_q) \right]}, \quad (4)$$

where $o$ is the total number of classes.

The individual probabilities on the right-hand side of this equation are estimated based on the weighted data. The prior probability for class $c_l$ becomes

$$p(c_l) = \frac{1 + \sum_{i=0}^{n} I(c_i = c_l) w'_i}{o + \sum_{i=0}^{n} w'_i}, \quad (5)$$

where $c_i$ is the class value of the training instance with index $i$, and the indicator function $I(x = y)$ is one if $x = y$ and zero otherwise.

Assuming attribute $j$ is nominal, the conditional probability of $a_j$ (the value of this attribute in the test instance) is given by

$$p(a_j|c_l) = \frac{1 + \sum_{i=0}^{n} I(a_j = a_{ij}) I(c_i = c_l) w'_i}{n_j + \sum_{i=0}^{n} I(a_j = a_{ij}) w'_i}, \quad (6)$$

where $n_j$ is the number of values for attribute $j$, and $a_{ij}$ is the value of attribute $j$ in instance $i$.

If the data contains a numeric attribute, we either discretize it using Fayyad and Irani's MDL-based discretization scheme (Fayyad & Irani, 1993), and treat the result as a nominal attribute, or we make the normality assumption, estimating the mean and the variance based on the weighted data. We will present empirical results for both approaches.

## 3 Experimental results

We first present some illustrative results on two artificial problems before discussing the performance of our method on standard benchmark datasets.

### 3.1 Evaluation on artificial data

In this section we compare the behaviour of locally weighted naive Bayes to that of the $k$-nearest neighbour algorithm on two artificially generated datasets. In particular, we are interested in how sensitive the two techniques are to the size of the neighbourhood, that is, the choice of $k$. We also discuss results for standard naive Bayes, using the normality assumption to fit the numeric attributes.

Figure 1 shows the first artificial dataset. This problem involves predicting which of two spheres an instance is contained within. The spheres are arranged so that the first sphere (radius 0.5) is completely contained within the larger (hollow) second sphere (radius 1.0). Instances are described in terms of their coordinates in three dimensional space. The dataset contains 500 randomly drawn instances from each of the two spheres (classes).

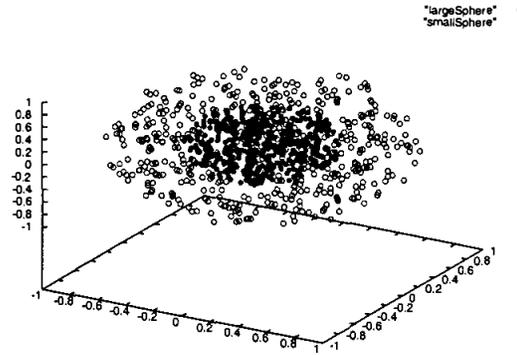

Figure 1: The two spheres dataset.

Figure 2 plots the performance of locally weighted naive Bayes (LWNB), $k$-nearest neighbours (KNN) and $k$-nearest neighbours with distance weighting[1] (KNNDW) on the two spheres data for increasing values of $k$. Each point on the graph represents the accuracy of a scheme averaged over the folds of a single run of 10-fold cross validation. From Figure 2 it can be seen that the performance of $k$-nearest neighbour suffers with increasing $k$ as more instances within an expanding band around the boundary between the spheres get misclassified. Locally weighted naive Bayes, on the other hand, initially improves performance up to $k = 40$ and then slightly decreases as $k$ increases further. The data is well suited to naive Bayes because the normal distributions placed over the dimensions within each sphere are sufficiently different. Standard naive Bayes achieves an accuracy of 97.9% on the two spheres data. When $k$ is set to include all the training instances, locally weighted naive Bayes gets 95.9% correct.

Figure 3 shows the second artificial dataset. This problem involves predicting whether an instance belongs to a black or white square on a checkers board given its $x$ and $y$ coordinates. 1000 instances were generated by randomly sampling values between 0 and 1 for $x$ and $y$. Each square on the checkers board has a width and height of 0.125.

Figure 4 plots the performance of locally weighted naive Bayes, $k$-nearest neighbours, and $k$-nearest neighbours with distance weighting on the checkers board data for increasing values of $k$. The strong interaction between the two attributes in this data makes it impossible for standard naive Bayes to learn the target concept. From Figure 4 it can be seen that locally weighted naive Bayes begins with very good performance at $k <= 5$ and then gracefully degrades to standard naive Bayes' performance of 50% correct by $k = 150$. In comparison, $k$-nearest neighbours' performance is far less predictable with respect to the value of $k$—it exhibits very good performance at $k <= 5$, quickly degrades to a minimum of 28% correct at $k = 60$, improves to 60% correct at $k = 150$ and then starts to decrease again.

---

[1] Our implementation of $k$-nearest neighbours with distance weighting uses the same linear weighting function as locally weighted naive Bayes (i.e. as given in Equation 2).

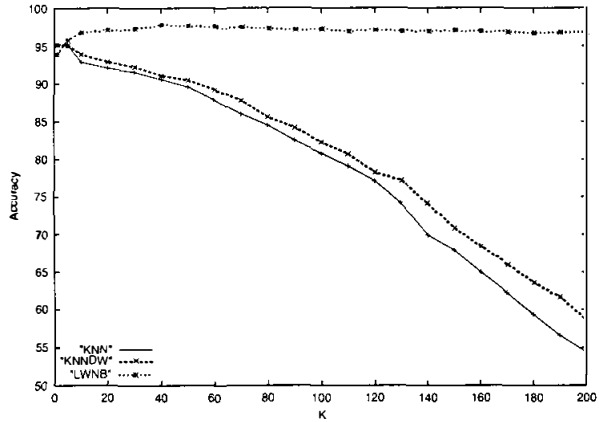

Figure 2: Performance of $k$-nearest neighbours (KNN), $k$-nearest neighbours with distance weighting (KNDW) and locally weighted naive Bayes (LWNB) on the two spheres data.

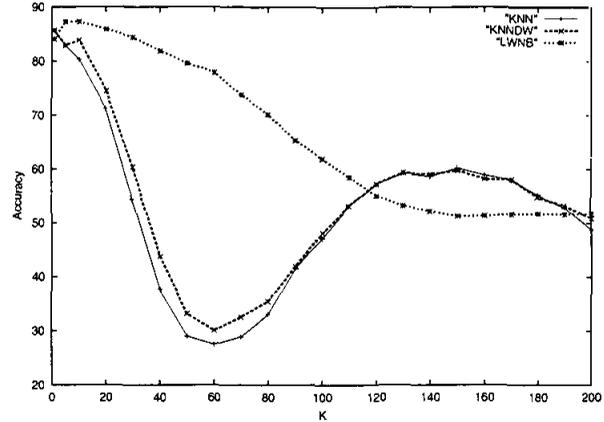

Figure 4: Performance of $k$-nearest neighbours (KNN), $k$-nearest neighbours with distance weighting (KNDW) and locally weighted naive Bayes (LWNB) on the checkers board data.

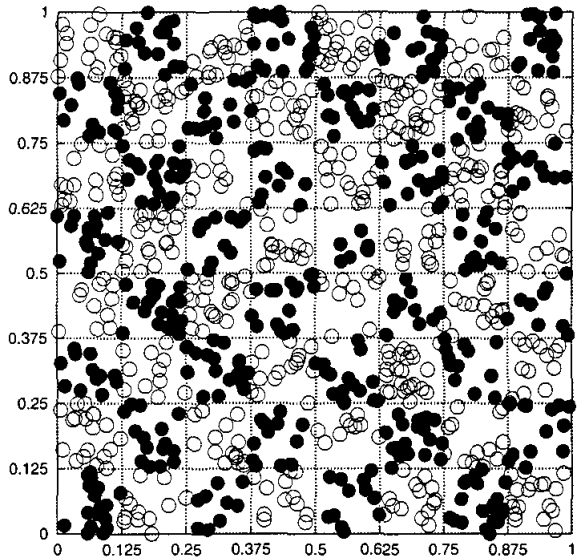

Figure 3: The checkers board dataset.

Table 1: Datasets used for the experiments

| Dataset | Inst. | % Msng | Num. | Nom. | Class |
|---|---|---|---|---|---|
| anneal | 898 | 0.0 | 6 | 32 | 5 |
| arrhythmia | 452 | 0.3 | 206 | 73 | 13 |
| audiology | 226 | 2.0 | 0 | 69 | 24 |
| australian | 690 | 0.6 | 6 | 9 | 2 |
| autos | 205 | 1.1 | 15 | 10 | 6 |
| bal-scale | 625 | 0.0 | 4 | 0 | 3 |
| breast-c | 286 | 0.3 | 0 | 9 | 2 |
| breast-w | 699 | 0.3 | 9 | 0 | 2 |
| diabetes | 768 | 0.0 | 8 | 0 | 2 |
| ecoli | 336 | 0.0 | 7 | 0 | 8 |
| german | 1000 | 0.0 | 7 | 13 | 2 |
| glass | 214 | 0.0 | 9 | 0 | 6 |
| heart-c | 303 | 0.2 | 6 | 7 | 2 |
| heart-h | 294 | 20.4 | 6 | 7 | 2 |
| heart-stat | 270 | 0.0 | 13 | 0 | 2 |
| hepatitis | 155 | 5.6 | 6 | 13 | 2 |
| horse-colic | 368 | 23.8 | 7 | 15 | 2 |
| hypothyroid | 3772 | 6.0 | 23 | 6 | 4 |
| ionosphere | 351 | 0.0 | 34 | 0 | 2 |
| iris | 150 | 0.0 | 4 | 0 | 3 |
| kr-vs-kp | 3196 | 0.0 | 0 | 36 | 2 |
| labor | 57 | 3.9 | 8 | 8 | 2 |
| lymph | 148 | 0.0 | 3 | 15 | 4 |
| mushroom | 8124 | 1.4 | 0 | 22 | 2 |
| optdigits | 5620 | 0.0 | 64 | 0 | 10 |
| pendigits | 10992 | 0.0 | 16 | 0 | 10 |
| prim-tumor | 339 | 3.9 | 0 | 17 | 21 |
| segment | 2310 | 0.0 | 19 | 0 | 7 |
| sick | 3772 | 6.0 | 23 | 6 | 2 |
| sonar | 208 | 0.0 | 60 | 0 | 2 |
| soybean | 683 | 9.8 | 0 | 35 | 19 |
| splice | 3190 | 0.0 | 0 | 61 | 3 |
| vehicle | 846 | 0.0 | 18 | 0 | 4 |
| vote | 435 | 5.6 | 0 | 16 | 2 |
| vowel | 990 | 0.0 | 10 | 3 | 11 |
| waveform | 5000 | 0.0 | 40 | 0 | 3 |
| zoo | 101 | 0.0 | 1 | 15 | 7 |

Table 2: Experimental results for locally weighted naive Bayes (LWNB) versus naive Bayes (NB) and $k$-nearest neighbours with distance weighting (KNNDW): percentage of correct classifications and standard deviation

| Data Set | LWNB $k=50$ | LWNB $k=30$ | LWNB $k=100$ | NB | KNNDW $k=5$ | KNNDW $k=10$ | KNNDW $k=20$ |
|---|---|---|---|---|---|---|---|
| anneal | 98.32±1.2 | 98.53±1.1 | 98.45±1.2 | 86.59±3.3 • | 97.32±1.6 | 96.28±1.7 • | 94.33±1.7 • |
| arrhythmia | 62.63±3.7 | 61.17±3.4 | 64.12±4.9 | 62.40±7.0 | 59.22±3.8 • | 59.45±2.8 • | 55.64±2.0 • |
| audiology | 78.89±6.7 | 77.97±6.8 | 78.23±6.4 | 72.64±6.1 • | 64.53±8.2 • | 58.42±7.2 • | 54.14±6.7 • |
| australian | 83.06±4.6 | 85.62±4.0 ○ | 79.55±4.4• | 77.86±4.2 • | 86.14±3.9 | 86.75±4.1 ○ | 85.71±4.2 |
| autos | 77.45±9.6 | 76.75±9.4 | 76.19±9.9 | 57.41±10.8• | 68.39±10.5• | 61.83±11.3• | 57.31±11.7• |
| bal-scale | 89.89±1.8 | 89.44±2.0 | 90.37±1.8 | 90.53±1.7 | 87.98±2.6 • | 90.27±1.9 | 90.14±1.8 |
| breast-c | 72.79±7.0 | 72.65±7.0 | 73.41±6.6 | 72.70±7.7 | 74.49±4.8 | 74.32±4.8 | 74.42±4.2 |
| breast-w | 96.28±2.2 | 96.01±2.4 | 96.72±2.1 | 96.07±2.2 | 97.01±2.0 | 96.81±2.1 | 96.54±2.2 |
| diabetes | 70.63±4.8 | 71.52±4.7 | 73.03±4.9○ | 75.75±5.3 ○ | 73.86±4.6 ○ | 73.75±4.5 ○ | 74.46±4.3 ○ |
| ecoli | 84.31±5.9 | 84.47±5.1 | 85.45±5.3 | 85.50±5.5 | 86.58±5.6 | 87.35±5.9 | 84.59±5.5 |
| german | 75.06±3.3 | 74.34±3.3 | 75.69±3.4 | 75.16±3.5 | 73.17±3.5 | 74.45±3.2 | 74.09±2.9 |
| glass | 72.35±8.3 | 70.61±8.2 | 69.64±8.8 | 49.45±9.5 • | 68.74±8.1 | 65.08±9.0 • | 62.73±8.5 • |
| heart-c | 81.42±6.1 | 82.38±6.3 | 80.47±7.4 | 83.34±7.2 | 82.13±6.2 | 82.19±6.1 | 82.59±6.3 |
| heart-h | 82.33±6.7 | 81.16±6.7 | 83.41±6.1 | 83.95±6.3 | 82.32±6.3 | 82.12±6.6 | 82.11±6.4 |
| heart-stat | 79.30±6.9 | 80.85±7.1 | 82.70±6.3 | 83.59±6.0 | 79.89±6.9 | 80.70±7.0 | 81.63±6.4 |
| hepatitis | 86.08±7.0 | 84.41±6.8 | 83.70±9.3 | 83.81±9.7 | 84.21±8.2 | 83.78±7.9 | 81.69±7.0 |
| horse-colic | 82.45±5.5 | 82.94±5.4 | 82.75±5.6 | 78.70±6.2 | 81.73±5.3 | 81.95±5.3 | 82.06±5.4 |
| hypothyroid | 96.39±0.9 | 95.72±0.8 • | 97.26±0.8○ | 95.30±0.7 • | 93.17±0.8 • | 93.18±0.7 • | 92.77±0.4 • |
| ionosphere | 83.30±4.7 | 80.23±5.0 • | 89.12±4.7○ | 82.17±6.1 | 85.10±4.7 | 84.27±4.9 | 83.48±4.8 |
| iris | 95.60±4.7 | 95.80±4.4 | 95.80±4.8 | 95.53±5.0 | 95.73±4.6 | 95.27±4.8 | 95.67±5.0 |
| kr-vs-kp | 97.78±0.8 | 97.76±0.8 | 97.77±0.7 | 87.79±1.9 • | 96.41±1.0 • | 95.54±1.2 • | 94.29±1.4 • |
| labor | 93.50±9.6 | 91.97±11.3 | 93.70±9.1 | 93.57±10.3 | 84.77±14.2 | 88.77±13.2 | 84.77±12.1 |
| lymph | 83.89±9.7 | 85.85±9.0 | 82.36±9.3 | 83.13±8.9 | 84.98±7.9 | 82.60±9.0 | 81.95±9.1 |
| mushroom | 100.0±0.0 | 100.0±0.0 | 100.0±0.0 | 95.76±0.7 • | 100.0±0.0 | 99.94±0.1 | 99.90±0.1 • |
| optdigits | 98.56±0.5 | 98.75±0.5 | 98.18±0.6• | 91.39±1.0 • | 98.73±0.5 | 98.69±0.5 | 98.26±0.6 |
| pendigits | 99.38±0.2 | 99.38±0.2 | 99.33±0.2 | 85.76±0.9 • | 99.27±0.2 | 99.10±0.3 • | 98.65±0.4 • |
| prim-tumor | 44.63±6.1 | 42.87±6.5 | 45.46±5.9 | 49.71±6.5 ○ | 46.43±6.8 | 46.70±6.4 | 46.37±5.8 |
| segment | 96.61±1.2 | 97.13±1.2 | 96.13±1.3 | 80.16±2.1 • | 95.50±1.3 • | 94.96±1.5 • | 94.10±1.6 • |
| sick | 96.82±0.7 | 96.77±0.7 | 96.80±0.7 | 92.75±1.4 • | 95.45±1.4 • | 95.57±1.3 • | 95.12±1.0 • |
| sonar | 88.00±5.9 | 87.70±6.9 | 89.05±6.2 | 67.71±8.7 • | 82.28±9.1 | 75.89±8.9 • | 70.94±8.6 • |
| soybean | 93.44±2.6 | 92.69±3.0 | 94.00±2.4 | 92.94±2.9 | 90.28±3.3 • | 88.07±3.2 • | 84.05±3.9 • |
| splice | 94.29±1.3 | 92.71±1.5 • | 94.97±1.2 | 95.41±1.2 ○ | 82.15±1.7 • | 85.10±1.7 • | 87.12±1.6 • |
| vehicle | 75.09±4.1 | 74.56±3.9 | 74.84±3.9 | 44.68±4.6 • | 71.49±4.1 • | 70.17±3.9 • | 68.68±3.9 • |
| vote | 95.38±2.8 | 95.03±3.0 | 96.20±2.6 | 90.02±3.9 • | 93.08±3.8 • | 92.92±3.7 • | 91.77±3.8 • |
| vowel | 95.59±2.4 | 95.65±2.4 | 98.34±1.3○ | 62.90±4.4 • | 93.86±2.8 | 72.56±5.8 • | 21.75±5.1 • |
| waveform | 81.88±1.8 | 81.79±1.9 | 82.34±1.8 | 80.01±1.4 • | 79.33±1.8 • | 81.12±2.0 | 83.12±1.9 ○ |
| zoo | 97.21±4.5 | 97.82±4.1 | 96.72±4.9 | 94.97±5.9 | 95.05±6.7 | 89.90±6.7 • | 87.15±6.5 • |

○, • statistically significant improvement or degradation over LWNB with $k = 50$

## 3.2 Evaluation on UCI datasets

This section evaluates the performance of locally weighted naive Bayes (LWNB) on a collection of 37 benchmark datasets from the UCI repository (Blake & Merz, 1998). The properties of these datasets are shown in Table 1.

We ran two experiments. The first compares locally weighted naive Bayes with $k = 50$ to standard naive Bayes (NB) and to $k$-nearest neighbours with distance weighting (KNNDW) using $k = 5, 10, 20$. Using distance weighting with $k$-nearest neighbours out-performed $k$-nearest neighbours without distance weighting on these datasets. We also ran locally weighted naive Bayes with $k = 30, 100$. In this experiment normal distributions were used by NB and LWNB for numeric attributes. The second experiment compares locally weighted naive Bayes to standard naive Bayes, a lazy Bayesian rule learner (LBR) (Zheng & Webb, 2000), tree augmented naive Bayes (TAN) (Friedman et al., 1997) and averaged one-dependence estimators (AODE) (Webb et al., 2003). In this case, since our implementations[2] of LBR, TAN and AODE can only handle nominal attributes, we discretized all numeric attributes using the method of Fayyad and Irani (1993).

All accuracy estimates were obtained by averaging the results from 10 separate runs of stratified 10-fold cross-

---

[2] All learning algorithms used in this paper, with the exception of TAN, are part of the freely available Weka machine learning workbench version 3.3.6 (http://www.cs.waikato.ac.nz/ml/weka).

Table 3: Experimental results for discretized locally weighted naive Bayes (LWNBD) versus discretized naive Bayes (NBD), lazy Bayesian rules (LBR), tree augmented naive Bayes (TAN) and averaged one-dependence estimators (AODE): percentage of correct classifications and standard deviation

| Data Set | LWNBD $k=50$ | LWNBD $k=30$ | LWNBD $k=100$ | NBD | LBR | TAN | AODE |
|---|---|---|---|---|---|---|---|
| anneal | 99.20±0.9 | 99.30±0.8 | 99.30±0.9 | 95.90±2.2 • | 98.01±1.5 • | 96.08±2.0 • | 97.75±1.5 • |
| arrhythmia | 69.36±4.2 | 69.45±4.3 | 70.51±4.6 | 72.04±5.5 | 73.88±5.4 ∘ | 69.85±5.4 | 72.50±5.4 ∘ |
| audiology | 78.89±6.7 | 77.97±6.8 | 78.23±6.4 | 72.64±6.1 • | 72.20±6.3 • | 76.00±8.4 | 72.28±6.2 • |
| australian | 85.06±3.7 | 84.91±3.9 | 85.13±3.7 | 86.22±3.8 | 86.10±3.9 | 85.65±3.9 | 86.75±3.8 |
| autos | 84.59±8.0 | 84.55±8.5 | 83.66±8.5 | 65.17±10.9• | 73.80±10.4• | 74.58±11.2• | 74.27±11.5• |
| bal-scale | 69.40±4.6 | 69.40±4.6 | 69.55±4.6 | 71.56±4.8 ∘ | 72.17±4.6 ∘ | 72.83±4.6 ∘ | 69.96±4.6 |
| breast-c | 72.79±7.0 | 72.65±7.0 | 73.41±6.6 | 72.70±7.7 | 72.35±7.8 | 71.34±7.5 | 72.57±7.2 |
| breast-w | 96.77±2.0 | 96.80±2.0 | 96.68±2.0 | 97.20±1.7 | 97.21±1.7 | 97.02±1.9 | 97.00±1.9 |
| diabetes | 74.44±4.6 | 74.17±4.7 | 74.79±4.7 | 75.26±4.8 | 75.38±4.7 | 75.52±5.0 | 75.70±4.7 |
| ecoli | 81.28±5.2 | 81.31±5.1 | 81.22±5.4 | 81.99±4.9 | 81.66±4.8 | 81.48±4.7 | 82.23±4.6 |
| german | 72.96±3.5 | 72.58±3.6 | 73.77±3.4 | 75.04±3.6 | 74.90±3.5 | 73.64±3.8 | 75.87±3.6 ∘ |
| glass | 74.50±9.7 | 74.28±10.0 | 74.41±9.3 | 71.79±8.9 | 72.22±8.8 | 71.45±8.2 | 74.39±8.3 |
| heart-c | 81.12±6.4 | 80.86±6.4 | 80.83±6.4 | 83.47±6.9 | 83.54±6.9 | 82.21±7.0 | 82.84±6.7 |
| heart-h | 82.81±6.6 | 82.20±6.5 | 83.37±6.2 | 84.20±6.3 | 84.54±6.3 | 84.29±6.4 | 84.10±6.3 |
| heart-stat | 83.63±6.1 | 83.85±6.0 | 83.37±6.4 | 82.56±6.1 | 82.59±6.1 | 82.85±6.7 | 82.70±6.6 |
| hepatitis | 84.82±7.9 | 84.27±7.7 | 85.16±9.0 | 84.28±10.3 | 84.91±9.7 | 86.26±8.6 | 85.22±9.2 |
| horse-colic | 82.61±5.5 | 82.56±5.4 | 82.88±5.8 | 79.54±5.8 | 82.33±5.8 | 80.87±5.8 | 82.99±5.6 |
| hypothyroid | 98.56±0.5 | 98.37±0.6 • | 98.76±0.5 | 98.19±0.7 | 99.12±0.5 ∘ | 99.02±0.5 ∘ | 98.56±0.6 |
| ionosphere | 92.42±4.3 | 92.48±4.5 | 91.88±4.0 | 89.29±5.0 • | 90.00±4.8 | 90.09±4.6 • | 91.06±4.7 |
| iris | 93.33±6.1 | 93.33±6.1 | 93.07±5.9 | 93.33±5.8 | 93.20±5.9 | 94.00 ±5.7 | 93.07±5.8 |
| kr-vs-kp | 97.78±0.8 | 97.76±0.8 | 97.77±0.7 | 87.79±1.9 • | 96.79±1.1 • | 94.32±1.3 • | 91.01±1.7 • |
| labor | 89.63±12.6 | 88.07±13.6 | 89.63±12.6 | 88.57±13.2 | 87.50±13.9 | 89.37±14.9 | 88.80±14.0 |
| lymph | 86.86±8.0 | 86.72±7.9 | 86.86±7.8 | 85.10±8.3 | 85.45±8.5 | 84.07±8.3 | 86.73±7.9 |
| mushroom | 100.0±0.0 | 100.0±0.0 | 100.0±0.0 | 95.76±0.7 • | 99.96±0.1 | 97.74±0.6 • | 99.97±0.1 |
| optdigits | 97.36±0.7 | 97.02±0.7 • | 97.78±0.6 ∘ | 92.17±1.0 • | 94.34±1.0 • | 95.61±0.8 • | 96.91±0.8 • |
| pendigits | 98.25±0.4 | 98.01±0.4 • | 98.50±0.4 ∘ | 87.72±1.0 • | 96.18±0.6 • | 94.37±0.7 • | 97.77±0.4 • |
| prim-tumor | 44.63±6.1 | 42.87±6.5 | 45.46±5.9 | 49.71±6.5 ∘ | 48.85±7.3 | 48.56±6.5 | 49.68±6.8 ∘ |
| segment | 95.77±1.3 | 95.41±1.2 | 96.28±1.2 ∘ | 91.16±1.7 • | 93.94±1.4 • | 91.34±1.7 • | 95.09±1.3 |
| sick | 97.47±0.7 | 97.43±0.8 | 97.40±0.7 | 97.12±0.8 | 97.66±0.8 | 97.42±0.8 | 97.36±0.8 |
| sonar | 76.06±9.6 | 76.01±9.7 | 76.07±9.7 | 76.23±9.5 | 76.04±9.7 | 75.41±10.1 | 76.56±9.5 |
| soybean | 93.44±2.6 | 92.69±3.0 | 94.00±2.4 | 92.94±2.9 | 93.41±2.7 | 92.34±3.0 | 93.41±2.8 |
| splice | 94.29±1.3 | 92.71±1.5 • | 94.97±1.2 | 95.41±1.2 ∘ | 95.80±1.1 ∘ | 95.60±1.1 ∘ | 96.07±1.0 ∘ |
| vehicle | 71.43±4.0 | 71.10±4.0 | 71.83±4.1 | 61.21±3.4 • | 69.53±3.9 | 62.86±3.6 • | 70.43±3.6 |
| vote | 95.38±2.8 | 95.03±3.0 | 96.20±2.6 | 90.02±3.9 • | 94.11±3.3 | 92.74±3.9 • | 94.34±3.4 |
| vowel | 87.14±3.4 | 87.31±3.2 | 86.69±3.7 | 58.56±5.3 • | 74.67±3.8 • | 64.59±5.1 • | 76.87±4.7 • |
| waveform | 82.00±1.7 | 80.81±1.7 • | 82.96±1.6 ∘ | 79.97±1.4 • | 83.42±1.6 ∘ | 82.01±1.5 | 85.00±1.5 ∘ |
| zoo | 96.25±5.6 | 96.25±5.6 | 96.05±5.6 | 93.21±7.3 | 93.21±7.3 | 94.85±6.6 | 94.66±6.4 |

∘, • statistically significant improvement or degradation over LWNBD with $k=50$

validation. In other words, each scheme was applied 100 times to generate an estimate for a particular dataset. In the case where discretization is applied as a pre-processing step, the intervals are first estimated from the training folds and then applied to the test folds. Throughout, we speak of two results for a dataset as being "significantly different" if the difference is statistically significant at the 5% level according to the corrected resampled $t$-test (Nadeau & Bengio, 1999), each pair of data points consisting of the estimates obtained in one of the 100 folds for the two learning schemes being compared. We also show standard deviations for the 100 results.

Table 2 shows the results for the first experiment. Compared to standard naive Bayes, locally weighted naive Bayes ($k=50$) is significantly more accurate on 17 datasets and significantly less accurate on only three datasets. In many cases our method improves the performance of naive Bayes considerably. For example, on the vowel data accuracy increases from 63% to 95.6%. Similar levels of improvement can be seen on glass, autos, pendigits, sonar, vehicle and segment. Compared to $k$-nearest neighbours, locally weighted naive Bayes ($k=50$) is significantly more accurate on 13 datasets when $k=5$ for $k$-nearest neighbours, 17 datasets when $k=10$, and 18 datasets when $k=20$. Locally weighted naive Bayes ($k=50$) is significantly less accurate than at least one of the three versions of $k$-nearest neighbours on diabetes, australian and waveform. Table 2 also shows that the accuracy of locally weighted naive Bayes is relatively insensitive to the value of $k$, although on some datasets there can be a significant difference.

Table 3 shows the results for the second experiment. This experiment compares discretized locally weighted naive Bayes to discretized naive Bayes, lazy Bayesian rules, tree augmented naive Bayes and averaged one-dependence estimators. When compared to naive Bayes, the locally weighted version ($k = 50$) is significantly more accurate on 13 datasets and significantly less accurate on three. Similar to the situation in the first experiment, many of the improvements over naive Bayes are quite considerable. When compared to lazy Bayesian rules, locally weighted naive Bayes ($k = 50$) is significantly better on eight datasets and significantly worse on five. Compared to tree augmented naive Bayes it is significantly better on 11 datasets and significantly worse on three. Against averaged one-dependence estimators, the result is seven significant wins in favour of locally weighted naive Bayes versus five significant losses. Table 3 also shows that there is very little difference in accuracy for different values of $k$ in locally weighted naive Bayes, although it appears that $k = 100$ produces slightly better results than $k = 50$.

## 4 Related work

There is of course a lot of prior work that has tried to improve the performance of naive Bayes. Usually these approaches address the main weakness in naive Bayes—the independence assumption—either explicitly by directly estimating dependencies, or implicitly by increasing the number of parameters that are estimated. Both approaches allow for a tighter fit of the training data.

Typically the independence assumption is relaxed in a way that still keeps the computational advantages of pure naive Bayes. Two such methods are tree-augmented naive Bayes (Friedman et al., 1997) and AODE (Webb et al., 2003). Both enable some attribute dependencies to be captured while still being computationally efficient.

Some alternative approaches try to transform the original problem to a form that allows for the correct treatment of some of the dependencies. Both semi-naive Bayes (Kononenko, 1991) and the Cartesian product method (Pazzani, 1996) are such transformation-based attempts for capturing pairwise dependencies.

Methods that implicitly increase the number of parameters estimated include NBTrees (Kohavi, 1996) and Lazy Bayesian Rules (Zheng & Webb, 2000). Both approaches fuse a standard rule-based learner with local naive Bayes models. The latter is similar to our approach in the sense that it is also a lazy technique, albeit with much higher computational requirements. Another technique is recursive naive Bayes (Langley, 1993), which builds up a hierarchy of naive Bayes models trying to accommodate concepts that need more complicated decision surfaces.

## 5 Conclusions

This paper has focused on an investigation of a locally-weighted version of the standard naive Bayes model similar in spirit to locally-weighted regression. Empirically, locally-weighted naive Bayes outperforms both standard naive Bayes as well as nearest-neighbor methods on most datasets used in this investigation. Additionally, the new method seems to exhibit rather robust behaviour in respect to its most important parameter, the neighbourhood size.

Considering the computational complexity, locally weighted naive Bayes' runtime is obviously dominated by the distance computation. Assuming a naive implementation of nearest neighbour this operation is linear in the number of training examples for each test instance. Improvements can be made by using more sophisticated data structures like KD-trees. As long as the size of the selected neighbourhood is either constant or at least a sublinear function of the training set size, naive Bayes could be replaced by a more complex learning method. Provided this more complex method scales linearly with the number of attributes this would not increase the overall computational complexity of the full learning process. Exploring general locally-weighted classification will be one direction for future work. Other directions include exploring different weighting kernels and the—preferably adaptive—setting of their respective parameters. Application-wise we plan to employ locally-weighted naive Bayes in text classification, an area where both standard naive Bayes and nearest-neighbor methods are quite competitive, but do not perform as well as support vector machines.

## 6 Acknowledgements

Many thanks to Zhihia Wang and Geoffrey Webb for kindly providing us with the TAN implementation.